\documentclass[conference]{IEEEtran}
\IEEEoverridecommandlockouts
\usepackage{cite}
\usepackage{amsmath,amssymb,amsfonts}
\usepackage{algorithmic}
\usepackage{graphicx}
\usepackage{textcomp}
\def\BibTeX{{\rm B\kern-.05em{\sc i\kern-.025em b}\kern-.08em
    T\kern-.1667em\lower.7ex\hbox{E}\kern-.125emX}}
\begin{document}
	
\title{Predicting the Flu from Instagram}

\author{Oguzhan~Gencoglu, Miikka~Ermes
\thanks{O. Gencoglu is with Faculty of Medicine and Health Technology, Tampere University, Tampere, 33014, Finland e-mail: (oguzhangencoglu90@gmail.com).}
\thanks{M. Ermes is with Tieto Ltd., P.O.Box 449, 33101, Tampere, Finland}
}

\maketitle

\begin{abstract}
Conventional surveillance systems for monitoring infectious diseases, such as influenza, face challenges due to shortage of skilled healthcare professionals, remoteness of communities and absence of communication infrastructures. Internet-based approaches for surveillance are appealing logistically as well as economically. Search engine queries and Twitter have been the primarily used data sources in such approaches. The aim of this study is to assess the predictive power of an alternative data source, Instagram. By using 317 weeks of publicly available data from Instagram, we trained several machine learning algorithms to both nowcast and forecast the number of official influenza-like illness incidents in Finland where population-wide official statistics about the weekly incidents are available. In addition to date and hashtag count features of online posts, we were able to utilize also the visual content of the posted images with the help of deep convolutional neural networks. Our best nowcasting model reached a mean absolute error of 11.33 incidents per week and a correlation coefficient of 0.963 on the test data. Forecasting models for predicting 1 week and 2 weeks ahead showed statistical significance as well by reaching correlation coefficients of 0.903 and 0.862, respectively. This study demonstrates how social media and in particular, digital photographs shared in them, can be a valuable source of information for the field of infodemiology.
\end{abstract}


\section{Introduction}
Infectious diseases concern public health. Some infectious diseases, even with small number of initial cases, can cause huge epidemics; therefore, continuous monitoring and early detection is important~\cite{world2009global,fukuda2009pandemic,world1999recommended}. However, identifying, diagnosing and reporting infectious diseases is still a challenge for many countries~\cite{world2009global}. Surveillance systems for communicable diseases aim to ensure that the diseases are monitored efficiently and effectively~\cite{sahal2009communicable}. During past 2 decades, expert groups have recommended that public health surveillance should be enhanced in order to provide early warnings of emerging infectious diseases, but the systems are still limited and fragmented with uneven global coverage~\cite{morse2012public,fukuda2009pandemic}. While studies have shown that both developed and developing countries face challenges with surveillance systems~\cite{sahal2009communicable}, affordable and logistically appealing internet-based methods could be especially valuable in the developing countries.

Seasonal influenza is one of the most common infectious diseases. It is an acute respiratory infection caused by influenza viruses, and it can spread easily from person to person~\cite{influenza2014fact}. Influenza continues to be a major cause of acute respiratory illnesses, increasing morbidity and mortality throughout the world, increasing financial costs and consequentially causing significant economic burden~\cite{treanor2015influenza}. While yearly epidemics affect all populations, hospitalization and death occur mainly among high-risk populations including individuals with chronic conditions, pregnant women, elderly, children aged 6-59 months and healthcare workers~\cite{influenza2014fact}. Estimates indicate that worldwide annual epidemics result in about 3 to 5 million cases of severe illnesses and about 290,000 to 650,000 deaths~\cite{influenza2014fact}. Influenza epidemics possess certain easily identifiable characteristics, which have allowed their identification throughout history. These characteristics include immense attack rates and explosive spread of the disease. Symptoms of influenza, such as cough and fever, are very characteristic as well~\cite{treanor2015influenza}. Financial and productivity losses are due to increased levels of work and school absenteeism and sudden peak in demand for healthcare services~\cite{influenza2014fact}.

Utilizing Internet data for estimating incidents of influenza and influenza-like illnesses (ILI) dates back to 2006, in which anonymous search engine queries were used for the task~\cite{eysenbach2006infodemiology}. Since then, recent developments in Internet technologies have provided novel ways to detect and even predict the outbreak of epidemics. In particular, Internet search engine queries~\cite{ginsberg2009detecting}, Twitter~\cite{achrekar2011predicting}, online blogs~\cite{ertem2018optimal}, electronic health records~\cite{lu2018accurate} or other sources such as Wikipedia article access logs~\cite{hickmann2015forecasting}, weather data~\cite{chakraborty2014forecasting}, restaurant table reservations~\cite{chakraborty2014forecasting} etc. have been used to develop models to monitor and/or forecast official influenza cases. So far, the proposed methods employ tabular, time-series and textual data to build the predictive models. Recent advancements in the field of machine learning, i.e., deep learning~\cite{lecun2015deep}, especially in computer vision and image understanding, enables one to build novel approaches that can utilize image data from social media for disease surveillance. One promising social media platform for this purpose is Instagram~\cite{instagramWeb,gauthier2015instagram}.

Instagram is a photo and video-sharing social network with 500 million daily active users as of 2018~\cite{smith2018numbers}. 35\% of adults in United States state that they follow Instagram online or on their cell phone~\cite{smith2018social}. That percentage is around 71\% for the age group 18-24~\cite{smith2018social}. Publicly available Instagram data has been successfully used for identification of predictive markers for depression~\cite{reece2017instagram}, potential drug interaction monitoring~\cite{correia2016monitoring} and extracting nutritional and calorific information of food posts~\cite{sharma2015measuring}. In this work, we propose a machine learning pipeline to nowcast (estimation of current week) as well as forecast official weekly ILI cases in Finland using publicly available data from Instagram. In addition to timestamp and textual features of the Instagram posts such as hashtags, we also utilize the visual content (i.e. images) in them with the help of deep neural network models. We train and compare the performance of 9 machine learning methods for the task. To the best of our knowledge, this is the first study to employ images in social media for forecasting the influenza epidemics.

\section{Related Work}

There has been a significant number of studies regarding monitoring and predicting ILI from the Internet data. Several machine learning and statistical estimation methods have been proposed to nowcast and forecast official ILI cases using various different data sources such as search engine queries, Twitter, online blogs, Wikipedia article access logs etc. The proposed metrics and consequently the reported results for evaluating and comparing these approaches differ as well. Extensive reviews of influenza forecasting studies include~\cite{nsoesie2014systematic,chretien2014influenza,alessa2018review}, the one by Alessa and Faezipour~\cite{alessa2018review} being the most recent one.

When it comes to data sources to infer the count of official ILI cases from, Internet search engine queries and Twitter have been the most prevalent data sources. The main reason is the possibility of implementing estimation models with high predictive power due to high number of users creating substantial amount of data in these platforms. Search engine queries, e.g., Google Trends, Baidu Index, Google Flu Trends,~\cite{eysenbach2006infodemiology,polgreen2008using,ginsberg2009detecting,yuan2013monitoring,chakraborty2014forecasting,domnich2015age,woo2016estimating,guo2017monitoring,xu2017forecasting,yang2017using,lu2018accurate,zhang2018using,liang2018forecasting}, Twitter~\cite{lampos2010flu,aramaki2011twitter,achrekar2011predicting,broniatowski2013national,kim2013use,chakraborty2014forecasting,woo2016estimating,volkova2017forecasting,zhang2017predicting,lee2017forecasting,zhang2017forecasting,lu2018accurate,ertem2018optimal}, online blogs~\cite{woo2016estimating,ertem2018optimal}, electronic health records~\cite{michiels2017influenza,yang2017using,lu2018accurate} or other sources such as Wikipedia articles access logs~\cite{generous2014global,mciver2014wikipedia,hickmann2015forecasting,ertem2018optimal}, weather data~\cite{chakraborty2014forecasting}, restaurant table reservations~\cite{chakraborty2014forecasting} etc. have been used as data sources for flu forecasting. Several studies combine more than one data source as well to enhance the predictive power of their models~\cite{chakraborty2014forecasting,bardak2015prediction,broniatowski2015using,yang2017using,ertem2018optimal,woo2016estimating}.

Proposed methods include conventional time-series forecasting techniques such as auto-regressive moving average (ARMA), auto-regressive integrated moving average (ARIMA) or variants of these~\cite{achrekar2011predicting,domnich2015age,broniatowski2015using,lu2018accurate,yang2017using,xu2017forecasting}. Different machine learning and statistical estimation techniques has been used as well, such as linear regression~\cite{polgreen2008using,ginsberg2009detecting,yuan2013monitoring,bardak2015prediction,kim2013use}, ridge regression~\cite{bardak2015prediction,guo2017monitoring}, least absolute shrinkage and selection operator (LASSO)~\cite{lampos2010flu,woo2016estimating,xu2017forecasting,guo2017monitoring,mciver2014wikipedia}, support vector machine (SVM) regression~\cite{aramaki2011twitter,zhang2017predicting,woo2016estimating,liang2018forecasting}, k-nearest neighbor (kNN) regression~\cite{chakraborty2014forecasting} and neural networks~\cite{lee2017forecasting,volkova2017forecasting,xu2017forecasting}. 

Most frequent metrics to evaluate the performance of the proposed predictive models are Mean Absolute Error (MAE), Root Mean Squared Error (RMSE), Mean Absolute Percent Error (MAPE) and Pearson's correlation (extensive review in~\cite{chretien2014influenza}). Due to different population sizes under examination, MAE and RMSE can not be used for an objective comparison between studies and MAPE can not be used if the ground truth data has zero values due to its mathematical definition. Overall, several studies report a Pearson's correlation coefficient greater than 0.90 on their nowcasting predictions on the test (hold-out) data~\cite{ginsberg2009detecting,lee2017forecasting,lu2018accurate,achrekar2011predicting,zhang2017predicting,woo2016estimating,bardak2015prediction,yang2017using}. 

Publicly available Instagram data has not been utilized for forecasting epidemics yet; however, data collected from Instagram has been used for several other health-related analysis. In~\cite{reece2017instagram}, Instagram data from 166 individuals has been analyzed for identification of predictive markers for depression. Correia et al. collected close to 7,000 Instagram user timelines to monitor potential drug interaction~\cite{correia2016monitoring}. Extracting nutritional and calorific information of food posts has also been performed~\cite{sharma2015measuring}. Similarly, 3 million food related posts shared on Instagram has been analyzed to depict dietary choices in food deserts - urban neighborhoods or rural towns characterized by poor access to healthy and affordable food~\cite{de2016characterizing}. Cherian et al. identified hashtags and searchable text phrases associated with codeine misuse by analyzing the content of 1,156 sequential Instagram posts over the course of 2 weeks~\cite{cherian2018representations}. To explore young women's smoking behaviors Cortese et al. analyzed Instagram posts, revealing dangerous trends counteracting public health efforts such as normalization of tobacco use~\cite{cortese2018smoking}. Similarly, content analysis of marijuana-related posts on Instagram revealed potential influence on social norms surrounding marijuana use~\cite{cavazos2016marijuana}. Findings of content analysis of alcohol-related Instagram posts showed that majority of such posts by youth depict alcohol in a positive social context~\cite{hendriks2018social}. Topic modeling analysis of 96,426 Instagram posts has been carried out for characterizing the health topics that are prominently discussed on Instagram~\cite{muralidhara2018healthy}. Shuai et al. compared Logistic Regression (LR), SVM, decision tree and their own tensor model to detect users with a social network mental disorder on Instagram and Facebook~\cite{shuai2016mining}. LR and SVM models have also been proposed for age group prediction of Instagram users~\cite{han2016teens}.

\section{Methods}
\subsection{Data Collection}
\subsubsection{Surveillance data}
In Finland, the National Infectious Diseases Register is run by the virology unit of the National Institute for Health and Welfare (THL). There are specific guidelines and notification templates compiled by THL for doctors, healthcare centers, hospital districts and laboratories for mandatory reporting of about 70 diseases such as influenza. If the conditions for notification are met, healthcare professionals are obliged to report infectious disease cases to the register within seven days. The original notification is canceled or complemented if the notification is discovered false later on. The official statistics of the registry are publicly accessible and can be used freely under the open data license~\cite{thlLicense}. 

For this study, weekly ILI incidents reported by public primary healthcare register in Finland between the dates 30 April 2012 and 27 May 2018 (in total of 317 weeks) were used. The data is publicly available and accessible~\cite{officialTHL}.

\subsubsection{Instagram data}
We identified 7 keywords in Finnish language to be searched from the hashtags of the Instagram posts, namely \textit{cough}, \textit{fever}, \textit{flu}, \textit{influenza}, \textit{muscle ache}, \textit{sick}, \textit{throat ache}. These keywords correspond to the most common symptoms of ILI and we hypothesized that they would be often used in social media posts associated with ILI. We collected publicly available Instagram posts containing at least one of these hashtags between the dates 30 April 2012 and 27 May 2018 (in total of 317 weeks). A Python crawler was employed to collect only the URL to the image, main text and the timestamp of the post for each post containing one of the hashtags. Other fields such as username, comments, location, number of likes etc. were not collected. Posts containing videos were excluded as well. In order to preserve privacy, instead of storing the images, we stored only the URL pointers to the images that are hosted on Instagram. Reading of the images and further processing such as feature extraction was done programmatically without storing the images even though all posts were publicly available. 

The number of Instagram posts containing a given hashtag can be examined from Table~\ref{table1}. The rows of the table are not mutually exclusive, i.e., a post may contain several of these hashtags at the same time. Consequently, number of posts in Table~\ref{table1} sum up to 22,257 while the total number of unique posts is 20,994. On the average, posts contains 10.9 words (non-hashtag) and 7.2 hashtags. Table~\ref{table1} also shows the average word (non-hashtag) and hashtags counts for posts containing a given hashtag.

\begin{table}[htbp]
\centering
  \caption{List of hashtags for searching Instagram posts, Number of posts, average word and hashtag counts}
  \begin{tabular}{p{1.5cm}p{1.6cm}p{1.6cm}p{0.80cm}p{0.95cm}}
     \hline
     \normalfont Hashtag & English \linebreak Translation & Number \linebreak of Posts & Average \linebreak Word \linebreak Count & Average \linebreak Hashtag \linebreak Count \\
     \hline
     \#ysk{\"a} & cough & 661 & 10.9 & 8.1 \\
     \#kuume & fever & 3,863 & 9.6 & 6.9  \\
     \#flunssa & flu & 14,251 & 11.1 & 7.1 \\
     \#influenssa & influenza & 970 & 14.5 & 6.2  \\
     \#lihaskipu & muscle ache & 101 & 20.2 & 7.7 \\
     \#kipe{\"a} & sick & 1,861 & 9.5 & 8.2  \\
     \#kurkkukipu & throat ache & 550 & 10.5 & 7.4  \\
     \hline
  \end{tabular}
  \label{table1}
\end{table}

\subsection{Feature Extraction}

\begin{figure*}[!ht]
      \centering
          \includegraphics[width=2.0\columnwidth,trim={20.5cm 13.5cm 25.8cm 19cm},clip]{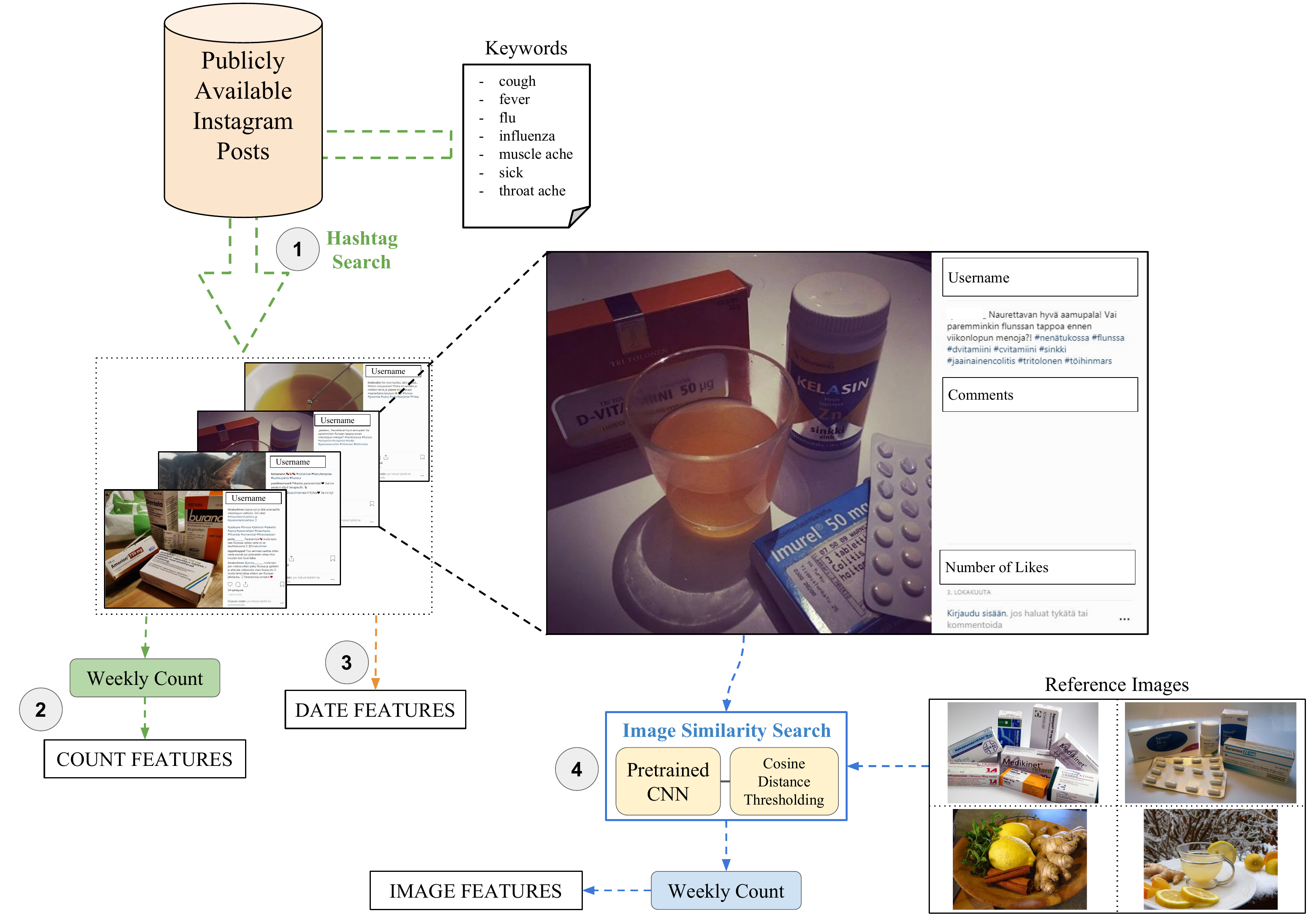}
      \caption{Proposed feature extraction scheme utilizing the date, text and image content of publicly available Instagram posts. 1 - search Instagram posts containing any of the selected keywords as hashtags, 2 - perform weekly counts of the collected posts for each keyword, 3 - extract the date features of the corresponding weeks, 4 - perform weekly counts of the posted images similar to reference images}
      \label{figure1}
\end{figure*}

Each week corresponds to a single observation, i.e., a single data point. We used the week number (between 1 and 52), month number (between 1 and 12) and the year, as numeric features. These 3 features will be referred as \textit{date features}.

For each hashtag (in total 7), we counted the weekly number of posts containing that particular hashtag, later referred to as \textit{count features}. Hashtags can be examined from Table~\ref{table1}.

\begin{figure}
      \centering
          \includegraphics[width=1.0\columnwidth,trim={0 13.1cm 0 0},clip]{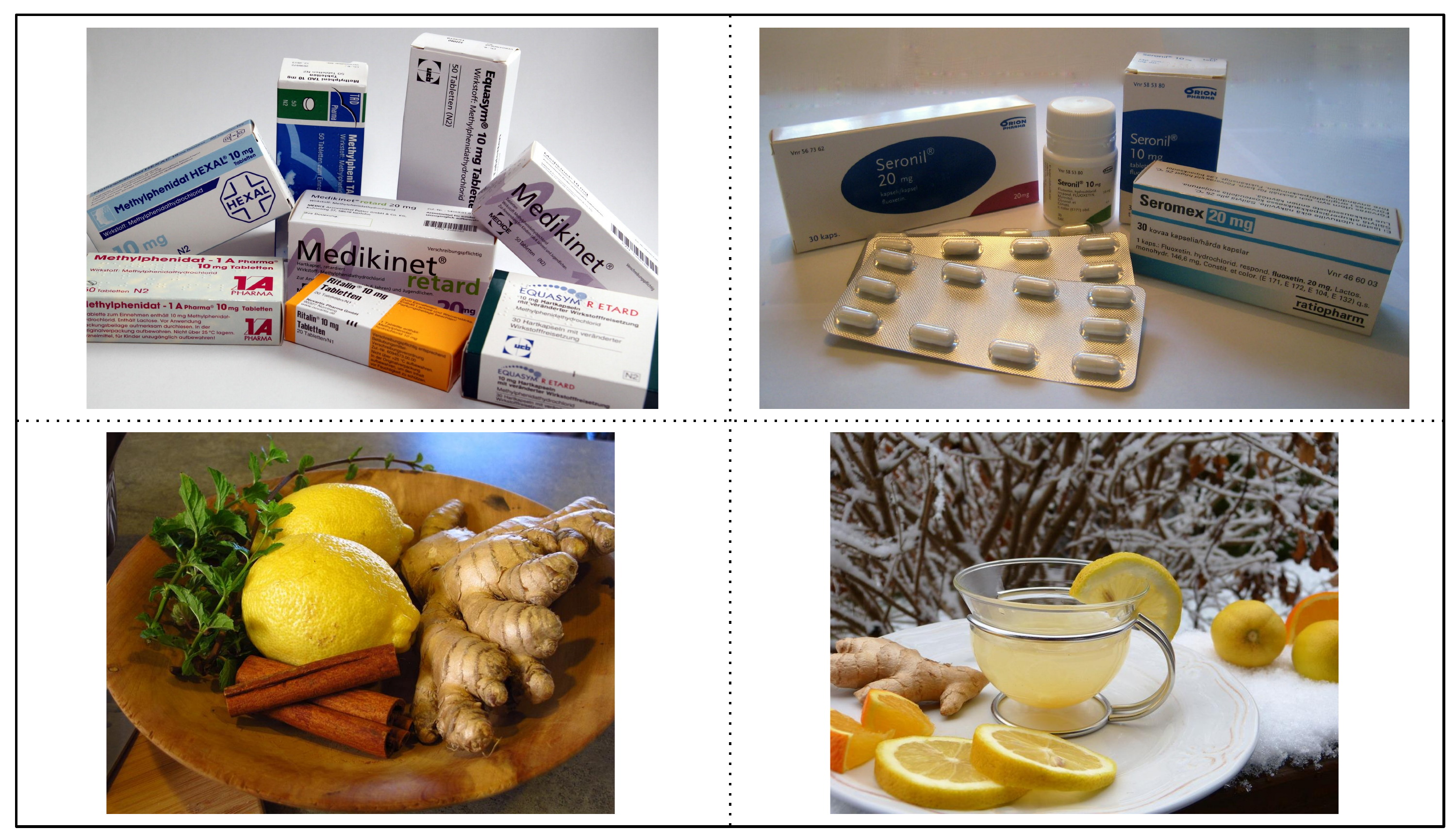}
      \caption{Selected 4 reference images for counting occurrence of similar images in Instagram posts.}
      \label{figure2}
\end{figure}

We selected 4 \textit{reference images} that contributed to the definition of 4 \textit{image features}. The images were collected using Google Images search engine and all of them are released under public domain enabling full permission for usage as is. The images were searched using terms "boxes of drugs/medicine", "boxes of drugs/medicine and pills", "mint, ginger and lemons" and "ginger and lemon tea", respectively~\cite{refimages} and they are shown collectively in Figure~\ref{figure2}. The search terms were defined by the authors' notion that images containing items like these were typically found in Instagram posts related to ILI. In order to count the weekly number of images similar to reference images on Instagram, we employed a pretrained deep convolutional neural network (CNN) model, i.e., Inception-ResNet-v2~\cite{szegedy2017inception}. This CNN architecture combines the two powerful architectures that has been proved to be successful for numerous computer vision tasks: Inception~\cite{szegedy2015going} and ResNet~\cite{he2016deep}. The model is 164 layers deep and has been pretrained on the well known ImageNet dataset~\cite{deng2009imagenet}. For each reference image, we obtained the vector representations out of the final layer before the fully-connected layers (an \textit{average pooling} layer), $V_{ref}^i \in \R$, resulting in a vector of length 1536 for each image. Similarly, we extracted the vector representations of each image collected from Instagram (from hashtag search), $V_{post}^j \in \R$. For each reference image vector, we computed the \textit{cosine distance}, $d_{ij}$ ($i=1,...,4, j=1,...,22257$), to every other Instagram image vector. 
\begin{equation}
d_{ij} = 1 - \frac{V_{ref}^i  \cdot V_{post}^j}{\left\lVert V_{ref}^i\right\rVert \left\lVert V_{post}^j\right\rVert}
\end{equation}
Cosine similarity/distance has been frequently used in content-based image retrieval tasks~\cite{bao2004comparative,babenko2015aggregating}. Smaller distance between the two representation vectors means those images are more likely to have similar visual content. For each reference image, we computed the \textit{mean}, $\mu_i$, and the \textit{standard deviation}, $\sigma_i$, of the distances (resulting in 4 mean and standard deviation pairs, one for each reference image). If the cosine distance between an Instagram image and a reference image is less then 2 unit standard deviations less from the mean, we incremented the weekly count corresponding to that reference image. This means, that particular Instagram image is very likely to have similar content to our reference image. Essentially, this is a \textit{similar image search} scheme and image features are simply the weekly counts of Instagram images similar to the selected reference images.

In total, concatenation of all features (date, count, and image) resulted in a feature vector of length 14. Technically, both count and image features are count-based features, i.e., count of Instagram posts by hashtags and count of images by similarity to reference images. A visual depiction of the proposed feature extraction scheme can be examined from Figure~\ref{figure1}. Each feature is normalized to zero mean and unity standard deviation (with respect to the training data) before regression modeling.

\subsection{Modeling}

We trained 9 different machine learning algorithms for nowcasting the official weekly ILI counts in Finland. These algorithms include \textit{linear regression} (also known as \textit{ordinary least squares}), \textit{ridge regression}, \textit{elastic net}, \textit{LASSO}, \textit{k-nearest neighbor regression}, \textit{support vector machine}, \textit{random forest}, \textit{AdaBoost} and \textit{XGBoost}. For each algorithm, we used date and count features for modeling. Furthermore, with XGBoost, we performed an extensive evaluation of the effect of different feature group combinations. XGBoost is a re-implementation of gradient boosting methods with enhanced regularization and speed functionality, gaining popularity since its announcement~\cite{chen2016xgboost}. XGBoost can also return the feature importances inherently which we reported. Additionally, we performed forecasting up to 3 weeks with the XGBoost model.

Implementation was done in Python (version 3.6) using \textit{scikit-learn}, \textit{xgboost} and \textit{TensorFlow} libraries~\cite{pedregosa2011scikit,abadi2016tensorflow,chen2016xgboost} on a 64-bit Ubuntu 16.04 workstation. All computations related to neural network models were performed on a single NVIDIA Titan Xp GPU. The rest of the machine learning algorithms were trained on a 20-core CPU in a parallel processing fashion to speed up the hyper-parameter search and cross-validation (CV).

\begin{table*}[!htbp] 
\centering
  \caption{Nowcasting results achieved by different machine learning models and corresponding input features}
  \begin{tabular}{p{3.70cm}p{3.2cm}p{1.25cm}p{1.7cm}p{0.8cm}p{0.7cm}p{2.1cm}}
     \hline
     \normalfont Model & Feature Extraction & Number of \linebreak Features & MAE \linebreak (10-fold~CV) & MAE \linebreak (test) & $R^{2}$ \linebreak (test) & Pearson's \linebreak Correlation (test) \\
     \hline
     1 - Elastic Net & date + count & 10 & \hspace{2 mm} 27.10 & 31.83 & 0.609 & \hspace{4 mm} 0.805 \\
     2 - XGBoost & date & 3 & \hspace{2 mm} 28.28 & 29.33 & 0.552 & \hspace{4 mm} 0.763 \\
     3 - XGBoost & count & 7 & \hspace{2 mm} 23.55 & 26.64 & 0.710 & \hspace{4 mm} 0.910 \\
     4 - SVM & date + count & 10 & \hspace{2 mm} 18.72 & 24.30 & 0.657 & \hspace{4 mm} 0.826 \\
     5 - AdaBoost & date + count & 10 & \hspace{2 mm} 18.11 & 18.35 & 0.781 & \hspace{4 mm} 0.915 \\
     6 - Random Forest & date + count & 10 & \hspace{2 mm} 17.04 & 17.61 & 0.861 & \hspace{4 mm} 0.938 \\
     7 - LASSO & date + count & 10 & \hspace{2 mm} 21.94 & 16.20 & 0.877 & \hspace{4 mm} 0.949 \\
     8 - Ridge Regression & date + count & 10 & \hspace{2 mm} 22.12 & 14.75 & 0.892 & \hspace{4 mm} 0.942 \\
     9 - Linear Regression & date + count & 10 & \hspace{2 mm} 22.55 & 14.66 & 0.886 & \hspace{4 mm} 0.943 \\
     10 - kNN Regression & date + count & 10 & \hspace{2 mm} 18.11 & 14.00 & 0.895 & \hspace{4 mm} 0.948 \\
     11 - XGBoost & date + count & 10 & \hspace{2 mm} 15.67 & 13.83 & 0.897 & \hspace{4 mm} 0.954 \\
     \textbf{12 - Deep ConvNet + XGBoost} & \textbf{date + count + image} & \textbf{14} & \hspace{2 mm} \textbf{13.14}  & \textbf{11.33} & \textbf{0.925} & \hspace{4 mm} \textbf{0.963} \\
     \hline
  \end{tabular}
  \label{table2}
\end{table*}

\subsection{Evaluation Scheme}
We used weekly data from 30 April 2012 to 22 May 2017 (265 weeks) as the training data, i.e., hyper-parameter optimization and model comparison. In order to report the performance of the trained models, data from one year was used as the test (hold-out) data, i.e., weekly data from 29 May 2017 to 27 May 2018 (52 weeks). The evaluation metric was chosen to be \textit{Mean Absolute Error} when comparing the models. To be able to successfully assess a machine learning model with a given set of hyper-parameters, we performed a 10-fold cross-validation on the training set. We ran our hyper-parameter search for each algorithm, calculating the average (over 10 folds) MAE. After best performing (achieving lowest MAE) hyper-parameters were set, we performed a final training of the model on the whole training set and evaluated the algorithms on the test data. Predictions of negative incidence counts were clipped to zero. We report MAE, coefficient of determination ($R^{2}$) and Pearson's correlation coefficient achieved on the test set.


\begin{figure*}[!htbp] 
      \centering
          \includegraphics[width=2.0\columnwidth,trim={0 0.4cm 0 0},clip]{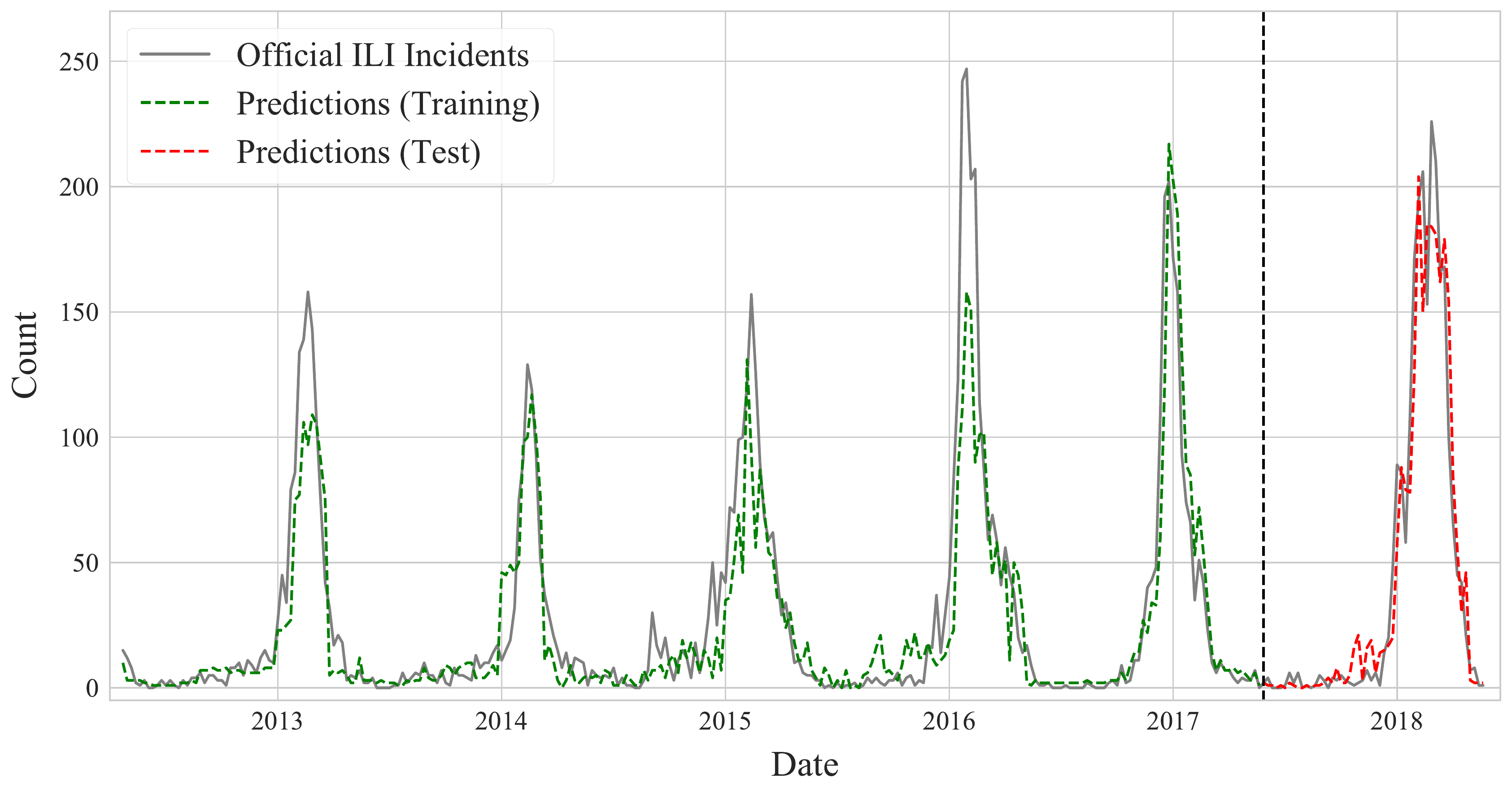}
      \caption{Weekly predictions of the best-performing nowcasting XGBoost model (model 12) and the official statistics.}
      \label{figure3}
\end{figure*}

\begin{figure}
      \centering
          \includegraphics[width=1.0\columnwidth,trim={0 0cm 0 0},clip]{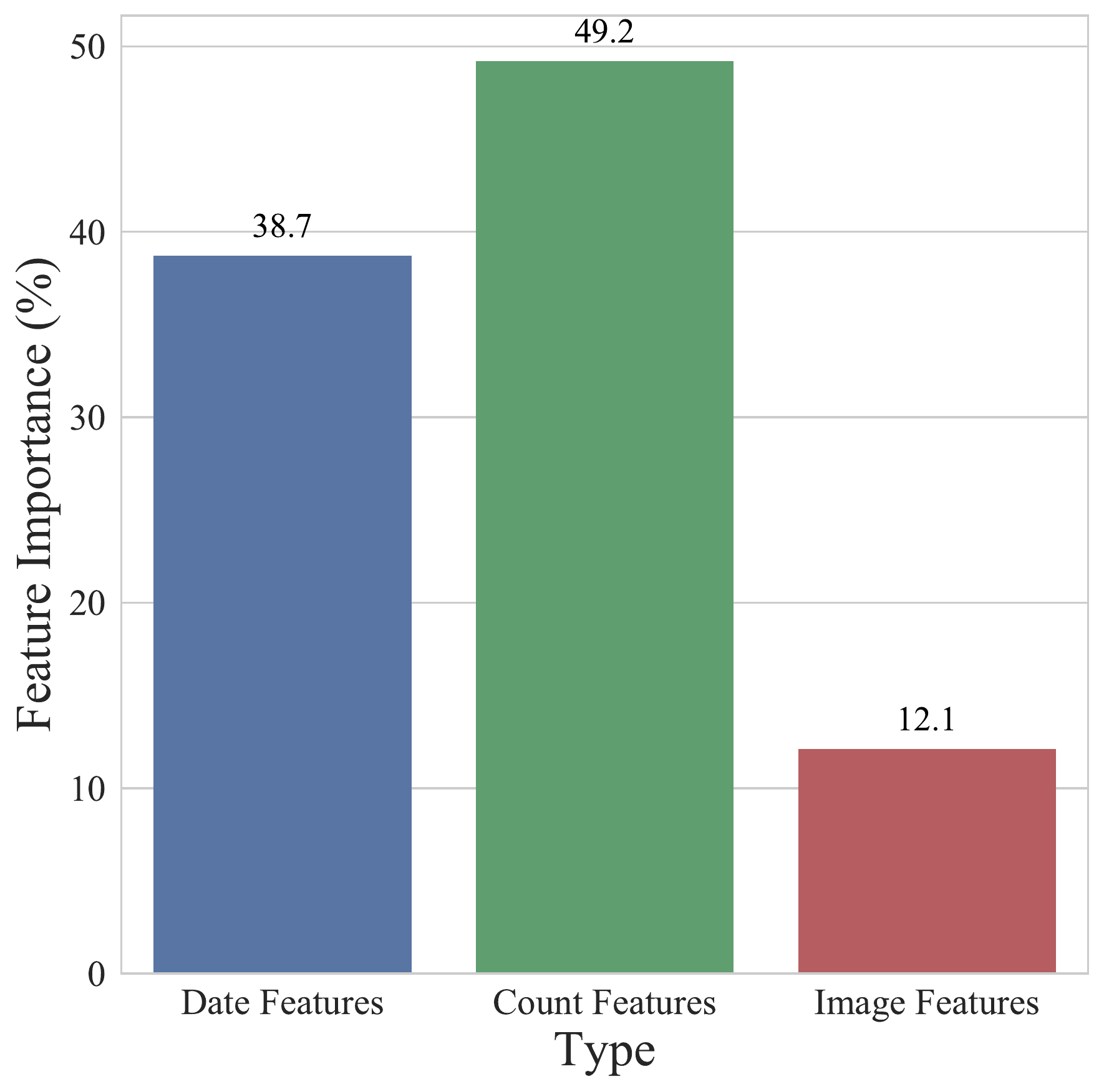}
      \caption{Relative feature importances of different feature modalities for nowcasting of ILI.}
      \label{figure4}
\end{figure}

\begin{table}[htbp]
\centering
  \caption{Performance of the forecasting models on the test data}
  \begin{tabular}{p{2.7cm}p{0.7cm}p{0.8cm}p{2.5cm}}
     \hline
     \normalfont & MAE & $R^{2}$ & Pearson's Correlation \\
     \hline
     nowcast (current week) & 11.33 & 0.925 & \hspace{6 mm} 0.963 \\
     1-week forecast & 17.84 & 0.814 & \hspace{6 mm} 0.903 \\
     2-weeks forecast & 23.38 & 0.687 & \hspace{6 mm} 0.862 \\
     3-weeks forecast & 35.54 & 0.205 & \hspace{6 mm} 0.685 \\
     \hline
  \end{tabular}
  \label{table3}
\end{table}

\section{Results}
Prediction results for nowcasting are shown in Table~\ref{table2}. The hyper-parameters that resulted in the lowest MAE on the 10-fold CV of the training data are as follows: regularization coefficient, $\alpha$, of 10.0 for ridge regression; regularization coefficient, $\alpha$, of 10.0 and $\ell_1\_ratio$ of 0.9, for elastic net; regularization coefficient, $\alpha$, of 1.0 for LASSO; number of neighbors of 6 and distance metric of \textit{euclidean} for kNN regression; regularization coefficient, $C$, of 100.0 and kernel of \textit{radial basis function} for SVM; number of estimators of 300 and maximum number of features of 3 for random forest; number of estimators of 300, loss function of \textit{linear} and learning rate of 0.001 for AdaBoost; number of estimators of 300, booster of type \textit{tree}, learning rate of 0.3 and $\ell_1$ regularization coefficient of 10.0 for XGBoost. $R^{2}$ scores and correlation coefficients were found to be statistically significant (\textit{p}\textless 0.001) for every model.

The official ILI statistics and the predictions of the best-performing nowcasting model, i.e., model number 12 involving XGBoost with date, count and image (extracted with a Deep ConvNet) features, can be seen in Figure~\ref{figure3}. As XGBoost is based on gradient tree boosting, it can inherently calculate the contribution of each feature to the decision. Such feature importance analysis results can be seen in Figure~\ref{figure4}. In addition, performance of the XGBoost model (model 12) on test data when predicting the future weeks, i.e, forecasting, can be examined in Table~\ref{table3}. $R^{2}$ scores and correlation coefficients were found to be statistically significant (\textit{p}\textless 0.001) for the forecasting models as well.

\section{Discussion}
Overall, we show that Instagram can be considered as a significant source of information for Internet-based monitoring and forecasting of influenza epidemics. Furthermore, we show that the visual content of the posted images can also be utilized as input features with the help of a deep convolutional neural network, increasing the prediction performance. A mean absolute error of 11.33 incidents per week and Pearson's correlation of 0.963 were achieved with XGBoost algorithm when several modalities of Instagram posts (date, count, image) have been used as an input for nowcasting the official influenza-like illness counts in Finland (see Table~\ref{table2}). The achieved MAE corresponds to 5.3\% of the average number of incidents observed in the highest 3 weeks of the test data, i.e., week 8, week 9 and week 6 of 2018 with 226, 210 and 206 incidents, respectively. For the forecasting model, accurate predictions up to 2 weeks were shown to be feasible as well with a correlation of 0.862 as shown in Table~\ref{table3}.

The proposed approach of image analysis can be utilized also for other social media platforms that contain images as part of their media. For instance, performance of communicable disease prediction models trained on Twitter data can be enhanced by incorporating content analysis of images in Twitter posts. Same idea can be extended to video content as well. Our results show that adding image features improves the prediction performance by decreasing the test MAE from 13.83 to 11.33. Even though the content of posted images on Instagram varies a lot, informative patterns for a given task can be filtered out with the proposed image similarity search. Our feature importance analysis show that the relative contribution of image features to the predictions corresponds to 12.1\% of all features (see Figure~\ref{figure4}). Considering only 4 reference images were used in this work to extract image features from, contribution of visual content of images to the predictive models can further be enhanced with an extensive set of reference images.

Robustness is a desired aspect of Internet-based disease surveillance systems. Different data sources can possess different robustness risks in terms of being a foundation for data-driven models. For instance, it was shown that in 2012-2013 flu season, Google Flu Trends failed to predict the number of flu incidents by estimating more than double the actual numbers~\cite{copeland2013google,butler2013google,lazer2014parable}. The cause was attributed to the heightened media attraction of the tool~\cite{copeland2013google}. Difficulty of replicability, therefore transparency, is another drawback of approaches based on search engine queries due to proprietary nature of the data. Social media platforms such as Twitter and Instagram possess robustness risks as well. Content analysis of Instagram images during the West African Ebola outbreak in 2014 showed that significant portion of the posted images were jokes or irrelevant to the outbreak itself~\cite{seltzer2015content}.

Our study was conducted by searching keywords in Finnish language and by using the influenza statistics of Finland as the reference. Focusing on a single language and country allowed us to fairly accurately identify the Instagram posts posted from this country while also obtaining accurate reference about the occurrence of influenza in the same country. Finnish language, unlike English, is spoken predominantly by Finnish people and people residing in Finland. Employing our approach in situations where a connection between a single language and country cannot be drawn (e.g. English-speaking countries) will require further filtering of posts by location which may not be available for all posts. In addition, only the publicly available Instagram data can be utilized for the proposed approach. Note that, public availability is a user-defined setting that can be changed anytime, rendering Instagram a dynamic source of information. This limitation holds true for other social media platforms such as Twitter or Facebook as well.

Future work will enable improvements by incorporating other input features derived from Instagram posts such as sentiment content of the post text and comments, location etc. The modeling performance can be enhanced with an extensive image similarity search as well as with ensembling several machine learning models together. Furthermore, normalization of the weekly counts with respect to the total number of weekly Instagram posts in the population under study (if available) may enhance the predictive performance by taking the popularity aspect of the platform into account. The ultimate goal is to implement robust and reliable predictive models that can use numerous data sources (Instagram, Twitter, Facebook, Wikipedia, search engines, blogs, weather data etc.) simultaneously to perform timely and accurate prediction of epidemics.

\section{Conclusion}

In summary, we show that Instagram is a valuable source of information for predicting influenza outbreaks. In addition to date and hashtag counts in publicly available Instagram posts, with the help of recent advancements in machine learning research, i.e., deep neural networks, we also utilized the image content of the posts in order to develop more accurate models for predicting influenza-like illnesses incidents in Finland. Future work includes validating our approach in wider scope geographically. We believe our work serves as an advancement in the field of infodemiology by highlighting a novel data source and advanced machine learning methods for enhanced accuracy and reliability of infectious disease surveillance.


\section*{Acknowledgment}

We thank Heidi Simil{\"a}, Marja Harjumaa and Mikko Virtanen who provided expertise that greatly assisted the research.

\bibliographystyle{IEEEtran}
\bibliography{references}

\end{document}